\documentclass[10pt,twocolumn,letterpaper]{article}

\usepackage{times}
\usepackage{iccv}
\usepackage{mwe}
\usepackage{epsfig}
\usepackage{graphicx}
\usepackage{amsmath}
\usepackage{amssymb}
\usepackage{booktabs}
\usepackage[ruled]{algorithm2e}
\usepackage{caption}
\usepackage[bookmarks=false]{hyperref}
\hypersetup{breaklinks=true,colorlinks}
\DeclareUnicodeCharacter{05BF}{d}

\begin{document}
\title{Semantify: Simplifying the Control of 3D Morphable Models using CLIP}
\author{
Omer Gralnik\\ \small Reichman University
\and 
Guy Gafni\\ \small Technical University of Munich
\and 
Ariel Shamir\\ \small Reichman University}
\date{}

\twocolumn[{%
\renewcommand\twocolumn[1][]{#1}%
\maketitle
\vspace{-8mm}
\begin{center}
    \centering
    \captionsetup{type=figure}
    \includegraphics[scale=0.76]{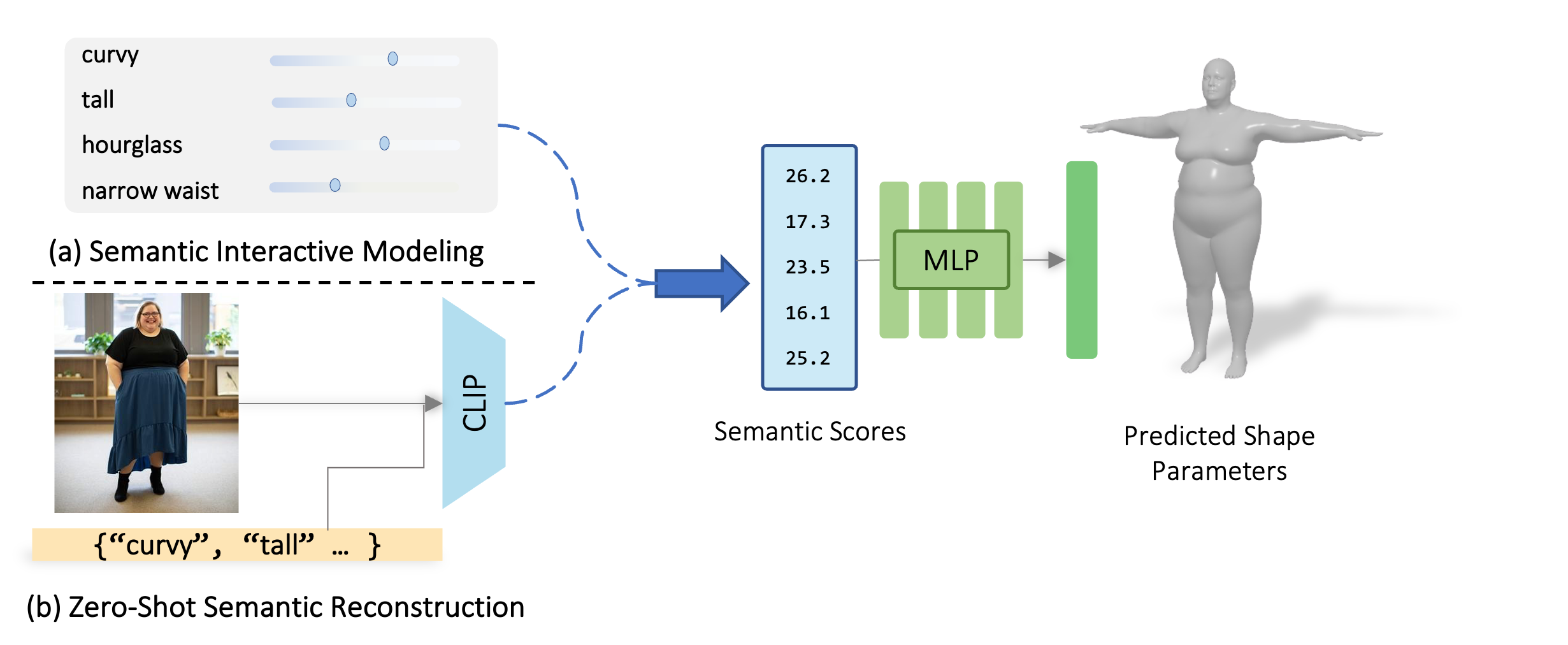}
    % \vspace{-4mm}
    \captionof{figure}{Semantify offers a method to create and edit a 3D parametric model using semantically meaningful descriptors. 
Semantify is based on a self-supervised method that utilizes the semantic power of CLIP language-vision model to build a mapping between semantic descriptors to 
3DMM model coefficients. This can be used in an interactive application defining a slider for each descriptor (a), or to fit a model to an image in a zero shot manner by feeding the image into CLIP and obtaining a vector of semantic scores that can be mapped to shape parameters (b).  
   }\label{fig:teaser}
\end{center}%
}]

\newcommand{\arik}[1]{\textcolor{red}{{#1}}}
\newcommand{\omer}[1]{\textcolor{blue}{{#1}}}
\newcommand{\guy}[1]{\textcolor{cyan}{{#1}}}
\newcommand{\as}[1]{\textcolor{red}{#1}}

\definecolor{todocolor}{RGB}{50,150,50}
\newcommand\TODO[1] {\emph{\textcolor{todocolor}{TODO: #1}}}

\vspace{-4mm}

\begin{abstract}
\vspace{-4mm}
We present Semantify: a self-supervised method that utilizes the semantic power of CLIP language-vision foundation model~\cite{radford2021learning} to simplify the control of 3D morphable models. 
Given a parametric model, training data is created by randomly sampling the model's parameters, creating various shapes and rendering them. The similarity between the output images and a set of word descriptors is calculated in CLIP's latent space. 
Our key idea is first to choose a small set of semantically meaningful and disentangled descriptors that characterize the 3DMM, and then learn a non-linear mapping from scores across this set to the parametric coefficients of the given 3DMM. The non-linear mapping is defined by training a neural network without a human-in-the-loop. 
We present results on numerous 3DMMs: body shape models, face shape and expression models, as well as animal shapes. We demonstrate how our method defines a simple slider interface for intuitive modeling, and show how the mapping can be used to instantly fit a 3D parametric body shape to in-the-wild images. See our project page at {\href{https://omergral.github.io/Semantify/}{https://omergral.github.io/Semantify/}}
\end{abstract}

\section{Introduction}
\label{sec:intro}

3D modeling techniques have evolved tremendously over the last few years. Such techniques are used in countless industries including the burgeoning AR/VR industry, fashion design, game development, film, and many others. However, designing a high-quality 3D model is not a simple task for most people and might require a well-trained 3D artist. Even when using more recent parametric morphable models (3DMM)~\cite{FLAME:SiggraphAsia2017,SMPL-X:2019,Zuffi:CVPR:2017,scape,xu2020ghum} it is still hard for humans to understand how to choose the correct set of parameters, e.g.\ to achieve a specific human body shape or facial expression. It is also difficult to find what are the limits of the given parametric model in terms of coverage and expressiveness. The reason is that in most cases, the provided set of parameters is not interpretable, as they are commonly calculated using automatic optimization mechanisms followed by dimensionality reduction using PCA. Thus, they carry no clear semantic meaning. 

To simplify the use of 3DMMs and allow for natural  interactive human modeling, a key research question is -- how to insert semantics into 3DMM control? Previous approaches~\cite{Streuber:SIGGRAPH:2016} relied mostly on human intelligence and labeling, which is often time consuming and expensive. In addition, previous methods create a large number of semantic control descriptors that are often correlated and entangled. This means it is difficult to anticipate the effect of each descriptor as changing a value in one may modify others, resulting in difficulties to control the model. 

In this paper, instead of using human labeling, we rely on the remarkable abilities of huge foundation models that combine natural language and visual understanding. 
We present a \emph{self-supervised} method that utilizes the semantic power of CLIP~\cite{radford2021learning} to define a method to control 3DMMs that carries two main advantages. First, it is more natural for humans to use as it allows modeling using a \emph{small} number of \emph{semantically meaningful} descriptors, that cover the space of deformations but are \emph{disentangled}. Second, it covers even extreme examples in the shape/pose space of parametric models as it utilizes CLIP's pre-training on a large number of images. The main idea is first to use CLIP to select a small subset of semantically meaningful and disentangled descriptors, and then learn a non-linear mapping of this set to the coefficients of the given 3DMM. 

\begin{figure*}
    \centering
    \includegraphics[scale=0.38,width=\textwidth]{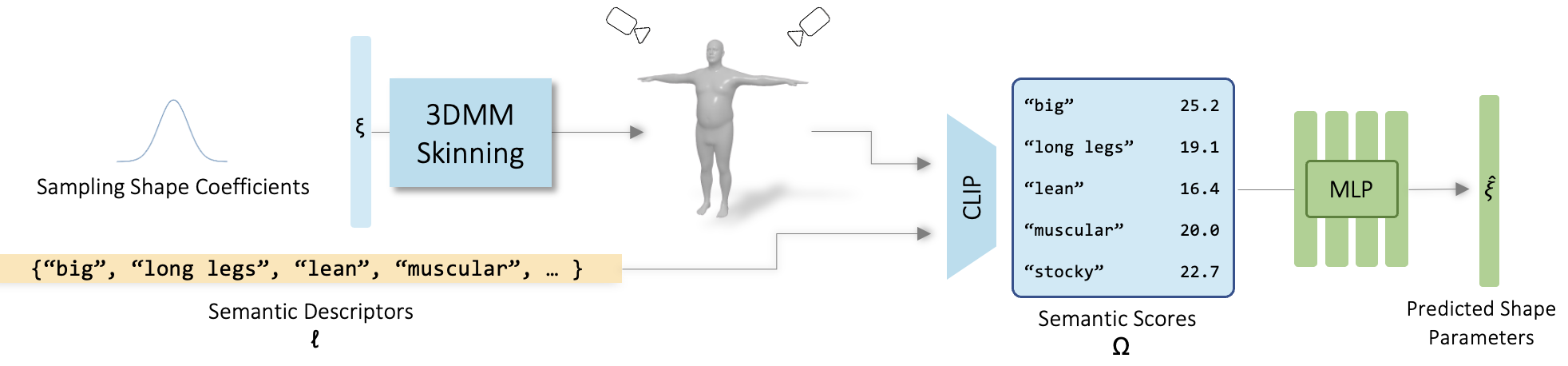}
    \caption{Learning a mapping from Semantic to Parametric space. (a) Given a coefficient vector ${\vec{\xi}}$ we create the 3DMM mesh. The mesh is rendered from several views. Each Rendered image ${I'}$ is passed into CLIP along with a set of semantic descriptors ${\vec{\ell}}$. The difference between each descriptor and the image in CLIP latent space is calculated and stored in the corresponding entry of the similarity vector ${\vec{\Omega}}$.  (b) Using a large set of such random pairs of $({\vec{\Omega}},{\vec{\xi}})$, we train a network to learn the mapping from semantic space to parametric space. }
    \label{fig:overview}
\end{figure*}

Given a parametric 3DMM, we sample its parameter space to create a dataset containing a variety of 3D mesh shapes. We then render each mesh from different camera views to create a diverse set of images corresponding to the parameter samples (see Figure~\ref{fig:overview}). 
Next, we gather a set of semantically descriptive textual terms related to the parametric 3D model, which we call \emph{descriptors}. We encode both the images (using CLIP's image encoder) and the descriptors (using CLIP's text encoder) into CLIP's latent space and compare them.
This defines a vector of similarity scores between the input vector of 3DMM coefficients (of each image sample) and each corresponding semantic descriptor.   
Next, we define a selection scheme to choose a small number of descriptors that are de-correlated to control the model. Lastly, we train a neural network to learn a mapping between the vector of similarity scores to the vector of 3DMM parameters. 

We demonstrate how the learned mapping can be used to define an interface to control a 3DMM in a way that is simple and effective using a small set of semantically meaningful sliders (see Figure~\ref{fig:teaser} and ~\ref{fig:sliders_app}). Such sliders are easy to employ for designing high-quality 3D models and cover the shape space well. We demonstrate this for four parametric models: human face's shape and expression (FLAME~\cite{FLAME:SiggraphAsia2017}), human body shapes (SMPL~\cite{smpl} and SMPL-X~\cite{SMPL-X:2019}), and even animals (SMAL~\cite{Zuffi:CVPR:2017}). We also show how the mapping can be used to instantly fit a 3D parametric body shape to an input image that works well ``in the wild'' even in extreme poses and body shapes. 

Our main contribution is a novel, self-supervised method for defining a small set of semantic descriptors to control a parametric model more naturally, by learning a mapping from a semantic space to a parametric representation without human-in-the-loop. We demonstrate the effectiveness of our approach on several parametric models and utilize it to define a simple interface for modeling and to instantly fit a 3D model to images in the wild. We will release our code for future research.

\section{Related Work}
\label{sec:related}

Our work encompasses a wide range of disciplines, including 3D parametric modeling, language-vision models, and zero-shot reconstruction.

\paragraph{3D Morphable Models.}

Much research has gone into improving 3D modeling of humans bodies, faces and other object classes, such as animals ~\cite{egger20203d,blanz1999morphable,smpl,SMPL-X:2019}. 3D Morphable Models (3DMM) are a powerful tool to parameterize the variation in geometry of objects belonging to a certain class. Dating back to 1999, Blantz et al.~\cite{blanz1999morphable} proposed 3DMMs to capture the variation of human faces and applied dimensionality reduction using Principal Component Analysis (PCA). 

In contrast to a 3D artist freely controlling the vertices of a mesh, using each of these PCA axes to deform the vertices induces a strong statistical prior on the resulting geometry. The resulting deformations are global in nature, strongly constraining the geometry to the manifold of statistically plausible meshes, as the per-axis coefficients correlate with their likelihood. While initially proposed for human faces, statistical 3DMMs have been developed for human body models \cite{scape,smpl,adam,SMPL-X:2019,xu2020ghum,supr}, facial models \cite{FLAME:SiggraphAsia2017,Tewari_2018_CVPR,wang2022faceverse,yang2020facescape}, and even 3D animal models \cite{Zuffi:CVPR:2017,8578514}. 

\paragraph{Semantic Models.}
PCA-based 3DMMs are efficiently and easily constructed, and admit an orthogonal and importance-ordered basis for the deformation. One disadvantage of these models is their lack of interpretability for humans. The axes of maximal variation in geometry, which form the deformation basis for each vertex, do not necessarily correspond to any semantically meaningful geometric change (e.g., raising an eyebrow), but instead, each axis induces a rather global change that may introduce correlations or effects that are unwanted to a human modeler, making it hard to obtain the desired shape or expression. 

Some 3DMMs \cite{bfm09_basel,2018BBaselV2} alleviate this issue by using deformation bases that are hand-crafted by artists, often referred to as \textit{blendshapes}. While this alleviates the interpretability issue, it loses the data-driven and explicit statistical prior, requires manual work, and does not guarantee the expressiveness of the resulting model. 

First to address the semantic issue of PCA-based 3DMMs in the human body domain were Seo et al.~\cite{Synthesizinga} which used metrics such as hip-to-waist ratio, body fat percentage, and height to regress 3DMM shape parameters. Hill et al.\cite{hill2015exploring} suggested examining the relationship between body shapes and description, by linking two similarity spaces - one created from body descriptions and the other from full-body laser scans. BodyTalk~\cite{Streuber:SIGGRAPH:2016} followed suit and approached this problem using the ``wisdom of the crowd''. In their work, they captured 256 male and female body shapes and represented each shape with 30 body descriptor words (e.g. curvy, fit, heavyset, round-apple, etc.). They hired labelers to discretely rate the bodies with respect to these descriptors and designed a regression model to learn the relationship between the discrete ratings and the parameters of the corresponding body shape.
In a more recent publication, SHAPY~\cite{Shapy:CVPR:2022} bridges this gap by curating datasets of images with corresponding body measurements (e.g., from model agencies) followed by human ratings respective to a set of linguistic shape attributes. They train models to predict SMPLX shape parameters from attributes and/or body measurements, and vice-versa.

\vspace{-0.5cm}
\paragraph{Language-Guided 3D Modeling.}
Recent works have shown great success in learning latent representations that are capable of coupling visual signals with language.
One such example is CLIP~\cite{radford2021learning} model. CLIP uses a contrastive training scheme \cite{contranstive} on over 400 million pairs of images and their captions to build a shared latent representation for visual-textual content. In this space, similarity between images and text can be computed. 
It has been shown that CLIP is an effective tool for image generation tasks. For example, providing natural-language guided manipulation of human face imagery in StyleGAN space \cite{patashnik2021styleclip,lou2022tecm,clip2styleGAN}.
Similarly, CLIP can be employed in 3D: using differentiable rendering, images of a 3D scene are rendered in a forward pass and are scored against a text prompt with CLIP. Taking the similarity score as an optimization objective, gradients are back-propagated through the CLIP network and the rendering process, back to the underlying  representation of the 3D scene. Text2Mesh~\cite{michel2022text2mesh} uses CLIP to optimize for colors and positions of mesh vertices to match a text query. CLIP-Mesh~\cite{Khalid2022CLIPMeshGT} followed suit and proposed to use CLIP to guide normal and texture maps. This basic recipe has been applied to a variety of scene representations, such as point clouds \cite{Nichol2022PointEAS} and Neural Fields~\cite{CLIP-NeRF,dreamFields,Wang2022NeRFArtTN,Gafni_2021_CVPR}. In the context of human faces and bodies, CLIP is used for generation of animation sequences \cite{Kim2022CLIPActorTR,tevet2022human,hong2022avatarclip,tevet2022motionclip,delmas2022posescript,aneja2022clipface}.

\vspace{-0.5cm}
\paragraph{Zero-Shot 3DMM Shape Reconstruction.}
Recovering accurate, explicit meshes from 2D signals such as images or video is an under-constrained and over-parameterized objective. The low-dimensional underlying representation of 3DMMs proves useful not only for modeling, but for 3D reconstruction of geometry from images of humans. The parameters can serve as a data-driven regularization term for shape and pose, and the low dimensional representation convexifies the optimization problem of fitting such a model to an image, rather than a freely deforming mesh.
Recent 3DMM-from-image methods either use iterative optimization schemes to fit the parameters of the 3DMM to the image \cite{kolotouros2019spin} or directly regress the shape and pose parameters from an Image ~\cite{PIXIE:3DV:2021,ExPose:2020,SPEC:ICCV:2021,kanazawaHMR18}. Our method allows to use CLIP image and text mapping and simply feed-forward the scores through an MLP network to get a body shape. 
\section{Method}
\label{sec:method}

An overview of our method can be seen in Figure~\ref{fig:overview}. In essence, our method defines a non-linear mapping from semantic space to parametric space by training a neural network to predict 3DMM coefficients from a vector of semantic attribute scores. We first create a dataset of rendered images of randomly sampled shapes (\ref{ssc:data-creation}), gather their corresponding semantic scores (\ref{ssc:clip-scores}). Then, we employ a scheme to reduce the semantic descriptors to a subset of least-correlated descriptors (\ref{ssc:Descriptor Selection}) and then train the network (\ref{ssc:training}).

\begin{table*}
\centering
\begin{tabular}{lccccc}
\toprule
\multicolumn{6}{c}{Coverage / Overlap}\\
\midrule
Method & Algorithm Choice & 2 descriptors & 5 descriptors & 10 descriptors & 15 descriptors \\
\midrule
 SMPLX-male & (6) / 92.1\% / 64.5\% & 38.6\% / 30.1\% & 51.5\% / 71.4\% & 49.7\% / 83.5\% & 98.3\% / 80.5\%\\
SMPLX-female & (4) / 49.7\% / 56.7\% & 27.1\% / 49.5\% & 50.0\% / 67.1\% & 47.4\% / 83.0\% & 82.4\% / 83.4\%\\
SMPLX-neutral & (8) / 52.0\% / 83.0\% & 40.2\% / 51.2\% & 50.0\% / 67.1\% & 51.9\% / 86.9\% & 52.1\% / 90.3\%\\
SMPL-male & (4) / 97.5\% / 60.2\% & 40.1\% / 46.7\% & 97.7\% / 66.5\% & 98.3\% / 83.9\% & 98.2\% / 89.3\%\\
SMPL-female & (5) / 95.7\% / 68.6\% & 36.3\% / 47.3\% & 95.7\% / 68.6\% & 97.8\% / 85.2\% & 99.1\% / 89.9\%\\
SMPL-neutral & (6) / 96.1\% / 76.0\% & 80.4\% / 48.7\% & 96.2\% / 71.5\% & 99.6\% / 85.2\% & 99.8\% / 90.8\%\\
FLAME-expression & (3) / 13.4\% / 56.8\% & 10.4\% / 47.4\% & 19.3\% / 67.2\% & 27.0\% / 77.3\% & 34.4\% / 85.0\% \\
\bottomrule
\end{tabular}
    \caption{Choosing different number of descriptors for a model. We evaluate the coverage and overlap of vertices for a mapper that was trained by using different number of descriptors (see Section~\ref{ssc:coverage}, the threshold used to measure coverage is 0.3). The number chosen by our algorithm is shown in parentheses at the beginning of each row. As can be seen, the more descriptors are chosen the larger the cover is, but the larger the overlap between the descriptors is as well. }
    \label{tab:coverage}
\end{table*}

\subsection{Dataset Creation} \label{ssc:data-creation}
\label{ssc:creating images}
The process of creating the data for training our model is similar across all 3DMMs (we demonstrate it on four different ones in this paper). For each  model, we only use the first 10 principal components, that cover above 95\% of the variance. We draw $N_{samples}$ 10-dimensional random vectors of coefficients. For body models, we draw a random shape coefficients vector $\vec{\beta}$, which yields a random shape. From the expression basis of the FLAME model, we randomly sample an expression coefficient vector $\vec{\psi}$ which yields a random expression. Since random values may also lead to noisy and unrealistic outputs, we limit the values of the sampled coefficients to a range of $k$ standard deviations of the model's coefficients ($k=2$ for body shapes and $k=4$ for face expressions), yielding more realistic results. The other parameters of the 3DMM $(\theta,\gamma)$ remain neutral.

Defining the semantic representation begins by gathering an over-complete set $\ell$ of $N_{\ell}$ word descriptors that correspond to each 3DMM. Our method supports any set of words that relate to the model: we couple the body model with body descriptors, and choose face descriptors for the face model. For the body models SMPL and SMPL-X, we adopt the set of descriptors used in \cite{Streuber:SIGGRAPH:2016}, for FLAME face expression we use an expanded set 
based on~\cite{EMOCA:CVPR:2022}, and for FLAME face shape we used our own set of descriptors. Lastly, since the SMAL model was trained on 5 animal families (Felidae-big cats, Canidae-dogs, Equidae-horses, Bovidae-cows, Hippopotamidae-hippos) we used a set of descriptors that contains animals from these given families.
The full list of descriptors can be found in the supplemental file. We note that our method can be used with any desired set of relevant descriptors.

\subsection{CLIP Ratings} 
\label{ssc:clip-scores}
We use CLIP~\cite{radford2021learning} to encode each text descriptor $\ell_{i}$ using their text encoder ${e}^{\ell_{i}} = {CLIP}_{Text}(\ell_{i})$. 
Each mesh created from a random coefficient vector ${F}(\vec{\beta},\theta,\gamma)$ is rendered to create a set of images ${I_j}$, and these images ${I_j}$ are encoded by CLIP image encoder ${e}^{I_j} = {CLIP}_{Image}(I_j)$ to the same latent space of the encoded text (see Figure~\ref{fig:overview}). In this space, we can compute the compatibility between the encoded image ${e}^{I_j}$ and each encoded label ${e}^{\ell_{i}}$ using cosine similarity to get a score $\Omega(I_j, \ell_{i}) = \cos({e}^{I_j},{e}^{\ell_{i}})$. This way, each random coefficients vector $\vec{\beta}$ or $\vec{\psi}$ is paired with a vector of the descriptors' scores:
$$\Omega_{i=1}^{N_{\ell}} = [\Omega(I_j, \ell_{1}), \ldots, \Omega(I_j, \ell_{N_{\ell}})] = [\Omega_{1}, ..., \Omega_{N_{\ell}}]$$ containing $N_{\ell}$ scores.

\subsection{Descriptors Selection} \label{ssc:Descriptor Selection}

There are many words that could describe a body shape, face shape or facial expression. As the score of a given image and word descriptor in CLIP's embedding space is calculated by their semantic similarity, there might be many possible word descriptors for a single image, and many possible images for a single word descriptor. For example, both the words ``happy'' and ``smile'' would have a high correlation with a smiling face, and therefore, their effect on a facial expression is entangled. Similarly, ``tall'' and ``short'' are entangled in a body shape model. Our key observation is that a large number of descriptors are difficult to handle for interactive modeling, and the larger the number, the more entangled they are in terms of their effect on the shape -- making it even harder to achieve the desired shape results (see Table~\ref{tab:coverage}). 

Contrary to BodyTalk~\cite{Streuber:SIGGRAPH:2016}, our goal is to use a minimal set of semantic word descriptors, while covering as much of the PCA shape space of each 3DMM as possible. 
We present an algorithm for selecting a subset of semantic descriptors with two main objectives: they should cover a large part of the shape space, and they should be disentangled as much as possible. 

Our idea is to first segment the shape space into regions, and assign descriptors to each region. This segmentation assures that we cover the shape space as we do not neglect any region. Next, we select for each region a set of descriptors by checking their variance (i.e., promoting coverage), and removing correlated descriptors, leaving the remaining ones as disentangled as possible. 

\textbf{Clustering.} To segment the shape space, we use our $N_{samples}$ dataset as a proxy to represent a sampling of the space and cluster their images (created as described in Section \ref{ssc:creating images}) encoded to CLIP's latent space.
we use K-means algorithm with Silhouette score to find $K$.

Next, we compare all encoded images $e^{I}$ in each cluster $C_{i}$ to all $N_{\ell}$ encoded word descriptors ${e}^{\ell_{i}}$ for $i=1,\ldots, N_{\ell}$. The top 5 most similar descriptors in a cluster are given a vote for this cluster. We normalize the value of votes by the size of each cluster. Then, each word descriptor is assigned exclusively to the cluster with the largest sum of votes for this descriptor. 

\textbf{Choosing descriptors.}
To anticipate the effect of a given descriptor on the 3DMM, we use the variance of its score $\Omega(I', \ell_{i})$ on all images $I'$ in the dataset. The larger the variance, the more descriptive this descriptor will be in terms of shape variation. To anticipate the level of entanglement between descriptors, we use their correlation. For each descriptor $\ell_{i}$ we build a vector of $N_{samples}$ scores $\Omega(I', \ell_{i})$ for all images $I'$ in the dataset. Correlation is measured between these vectors. The larger the correlation between two descriptors, the more similar the effect of this descriptor is on the shape, and the more entangled they are (we show some correlation plot examples in the supplemental file). Lastly, we apply antonyms and synonyms detection to verify there are no such pairs in our final set of descriptors.

The process of choosing the descriptors is first performed on each cluster separately, and then merging the lists. We sort the cluster descriptors according to their variance in descending order. Thus, the first descriptor that is chosen from each cluster is the descriptor with the highest variance. Once a descriptor is chosen, we iterate over the other descriptors that are left for this cluster and filter out the ones that have a high correlation with the chosen descriptor (we use the median correlation value of all pairs as the threshold). In addition, before adding the next descriptor to the set, we check if its synonyms or antonyms are already chosen, and if so, we skip it. 
This process continues until the list of descriptors is exhausted. 

Lastly, we merge the lists created from all clusters and sort again according to the descriptor variance. In a similar manner as we did for each cluster, we filter from the merged list all correlated descriptors, synonyms and antonyms. Finally, we arrive at a set of $d$ chosen semantic descriptors for this model. $d$ can be different for each model, but our algorithm can also support any preset number of descriptors by removing descriptors from the bottom of the list or adding back descriptors that were filtered out according to their variance order (see Table~\ref{tab:coverage}). 

Our algorithm can also support a preset list of descriptors if the user seeks to fit a 3D model with respect to some specific descriptors. In this case, the process described above for choosing the final set of descriptors is simply initialized with the user pre-defined descriptors, so that they are always contained in the final set.

\subsection{Training} \label{ssc:training}
For simplicity, we will denote the coefficients vector by $\vec{\xi}$ for both shape and expression parameters. Our goal is to define the mapping from the semantic representation of the $d$ descriptors to the parametric representation represented by $\vec{\xi}$.
From our dataset creation (Section \ref{ssc:creating images}), each coefficients vector $\vec{\xi} = [\xi_{1},...,\xi_{10}]$ is paired with a scores vector $\vec{\Omega} = [\Omega_{1}, ..., \Omega_{N_{\ell}}]$.
Hence, for a given scores vector $\vec{\Omega}$, the goal is to predict the corresponding coefficients vector $\vec{\xi}$. This mapping resembles the one presented in~\cite{Streuber:SIGGRAPH:2016}, only we rely on CLIP score to label our data, providing a single score rather than multiple scores obtained by crowd sourcing, and we do not assume a linear relationship between the word descriptors and the 3DMM's coefficients vector. Instead, we train a multi-layer-perceptron with ReLU activations as the mapping function using our paired data and a simple $L_2$ loss $\mathcal{L} = ||\hat{\vec{\xi}} - \vec{\xi}||_{2}$. The network consists of hidden layers (500 and 800 neurons), and is trained for 50 epochs.

\begin{figure}
\begin{center}
\includegraphics[width=0.5\textwidth]{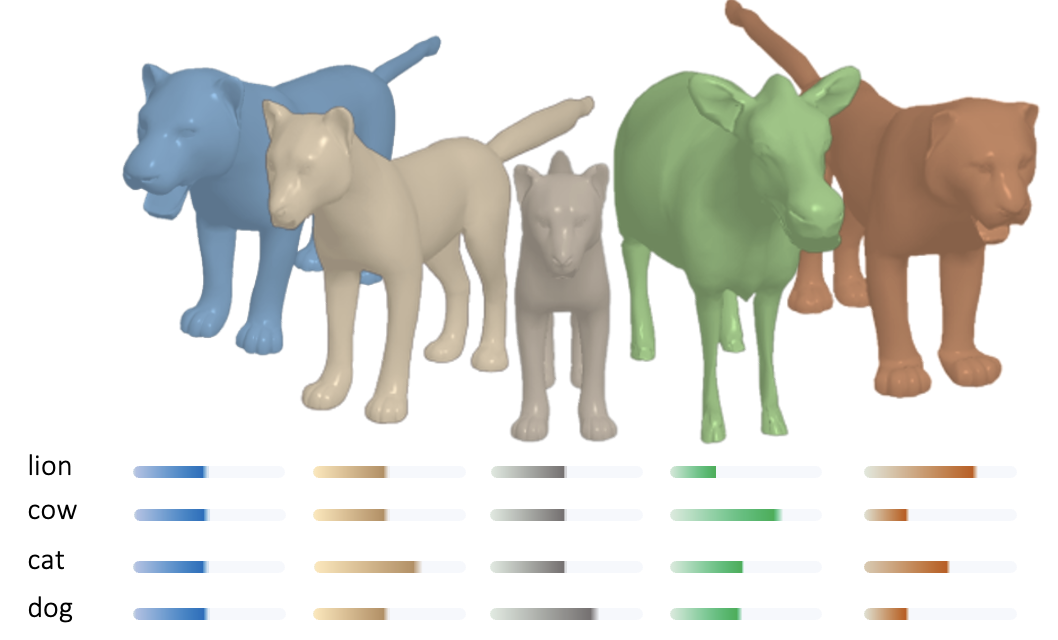}
    \caption{Models created using an interactive slider application for SMAL\cite{Zuffi:CVPR:2017} model (the blue model on the left is the neutral model). These  animals were created by interpolating between four semantic descriptors - lion, cow, cat and dog. As can be seen, other animals could also be designed as well as intermixed creatures. }
    \label{fig:smal}
\end{center}
\end{figure}

\section{Applications}

\subsection{Interactive Sliders}
\label{subsec:interactive_sliders}
\begin{figure}
\begin{center}
\includegraphics[scale=0.35]{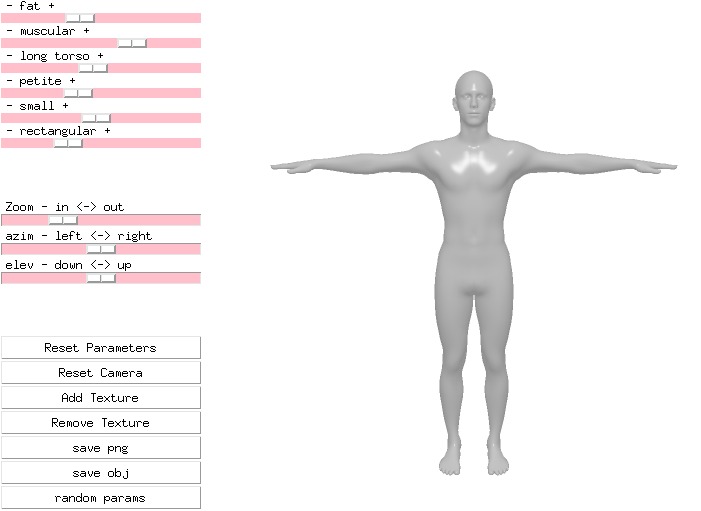}
    \caption{A simple and intuitive interactive application can be defined using sliders for each semantic descriptor. With Semantiy, the sliders have strong semantic meaning and are disentangled so that changing one will have very limited affect on others.}
    \label{fig:sliders_app}
\end{center}
\end{figure}

Using the set of semantic descriptors $\ell$ chosen by our algorithm, we can build a simple and intuitive interface to control the 3DMM. This is because the descriptors are least entangled and cover the PCA space well.
First, we train a mapper $M$ with the chosen set of descriptors $\ell$ as described above (Section~\ref{ssc:training}). Next, we create a slider for each descriptor whose value represents the desired CLIP score for that specific descriptor (see Figure~\ref{fig:sliders_app}).
The user can now change each slider to trigger a re-prediction of the 3DMM coefficients. The computation is interactive as the interface takes all sliders' values as input $\vec{\Omega}$ to $M$, which maps them in a forward pass to the 3DMM coefficients vector $\vec{\xi}$ and renders the corresponding 3D mesh. Thus, the modeler can interact and understand the relationships between the descriptors and their effect on the mesh. Figure~\ref{fig:smal} demonstrates examples of 3D meshes of different animals that were created using an interface created for the SMAL model.

\subsection{Zero-Shot Image to Shape Reconstruction}
Our method allows to leverage CLIP's semantic understanding to define a zero-shot image to shape reconstruction. 
Given any image of a person, we embed it into CLIP's latent space and use the similarity scores of the image against the set of descriptors $\ell^{M}$ chosen for a mapper $M$ to obtain the vector $\vec{\Omega}$. Next, we simply feed  $\vec{\Omega}$ through the network to get the 3DMM coefficients vector $\vec{\xi}$ that best fits the image and create the shape from $\vec{\xi}$. 
Our method differs from state-of-the-art methods in that it enables the user to refine the zero-shot prediction very simply using semantic sliders starting from a relatively good initial guess.

\section{Experiments}
\label{sec:experiments}

\subsection{Coverage Evaluation}
\label{ssc:coverage}

To evaluate a mapper $M$ that was trained with a given set of descriptors $\ell$, we would like to measure its effect on the 3DMM.
For each descriptor $\ell_{i} \in \ell$ we do a forward pass through the mapper twice: the first pass by setting a low score to $\Omega_{i}$ and the second pass by setting a high score to $\Omega_{i}$. The other coefficients are set to a default value and remain constant. Each such pass creates a parameter vector $\vec{\xi} = [\xi_{1},...,\xi_{10}]$, which represents the resulting 3DMM shape or expression. We denote these vectors as  $\vec{\xi}(i)_{low}$ and $\vec{\xi}(i)_{high}$.

\begin{figure}[h]
\begin{center}
\includegraphics[width=0.5\textwidth]{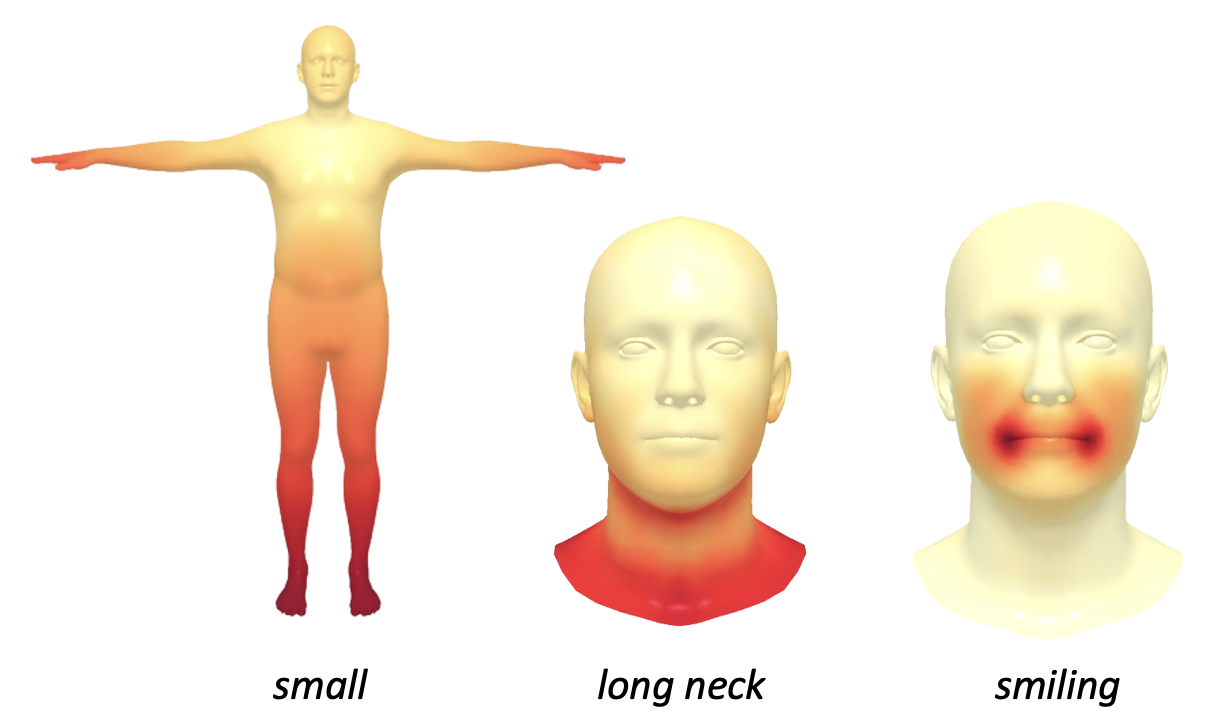}
    \caption{The geometric effect of a given descriptor. The darker the color of the vertex is (more red), the more it is affected by the descriptor. To define coverage we collect all vertices with a value larger than a given threshold $\theta$. }
    \label{vertex_coverage}
\end{center}
\end{figure}

We measure the geometric effect of a descriptor $\ell_{i}$ on a 3DMM by examining the deformation of the vertices of the mesh. This is evaluated by comparing the position of each vertex in the two extreme cases of $\vec{\xi}(i)_{low}$ and $\vec{\xi}(i)_{high}$. Hence, for each vertex $v$ the size of deformation is defined as
$\delta'(v) = ||v_{low} - v_{high}||_{2}$. We can normalize these values by $\delta_{\max} = \max_{v \in Mesh} \delta'(v)$ and get a value $\delta(v) = \delta'(v)/\delta_{\max}$ between 0 and 1 (see Figure~\ref{vertex_coverage}). 

\vspace{0.2cm}

Using these differences, we can evaluate the coverage of a specific descriptor by measuring the number and positions of vertices it affects in the mesh. For example, we can look at all vertices $v$ whose value is above a threshold $\delta(v) > \theta$ for any given descriptor. We can also compare the overlap between two descriptors by measuring the intersection-over-union of their covered vertices.
We use these measure in our ablation studies as described next in Section~\ref{ssc:ablations}.
Note that this measure is only an approximation as it measures the size of the change of vertices and not their direction. 

\subsection{Ablations}
\label{ssc:ablations}

\begin{figure*}
\centering
\includegraphics[width=0.95\textwidth]{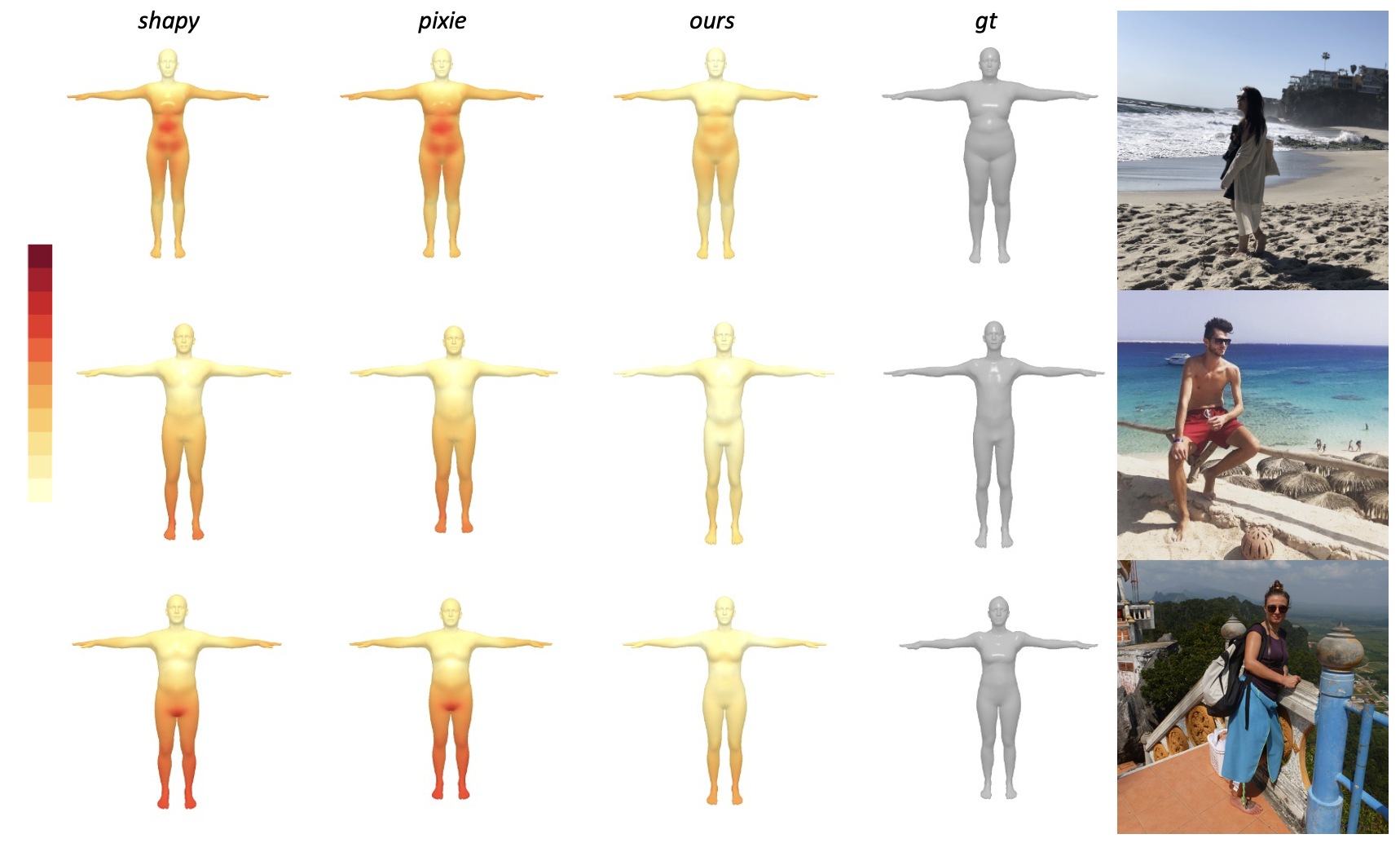}
    \caption{Results of zero-shot image to shape reconstruction. We compare Semantify against SHAPY and PIXIE on SMPLX 3DMM. Colors indicate error compared to ground truth (darker means larger error).}
    \label{fig:image2shape}
\end{figure*}

To further evaluate our algorithm's performance, we conducted experiments that include training the mapper with different numbers of descriptors and training on different sizes of training data.
As we do not have valid ground truth to estimate the accuracy of our mapper we created a test set that contains several body shapes that were generated by the first 10 blendshapes $\vec{\xi}$ of the SMPL-X model. In each experiment we optimize the descriptors' values $\vec{\Omega}$, to minimize the error $||\hat{\vec{\xi}} - \vec{\xi}||_{2}$. The goal of these experiments is to test the expressive power of the given configuration to match a given shape. The smaller the error the more expressive the method is.   
We show the results of these ablations in the supplemental file. We used this evaluation to choose the hyper-parameters we use in our algorithm: training with 3K sample images with texture for all models.

\vspace{0.3cm}
\subsection{Interactive Sliders User Study}
\label{ssc:sliders}

To evaluate Semantify's performance we conducted a user study that compares our approach against BodyTalk~\cite{Streuber:SIGGRAPH:2016}, as well as the original basis of the 3DMM. The user is presented with a  target 3D model (randomly created using the 3DMM) and is tasked to fit a 3DMM mesh model to the target shape using a set of sliders, using the two compared interfaces (see Figure~\ref{fig:sliders_app}) at random order. Users were given up to 5 minutes to complete the task.
BodyTalk was implemented on the SMPL body model, therefore we compare it to our SMPL male and female models. 
In another study, we compare our SMPLX male and female models to the original PCA-based axes. In both studies, we seek to evaluate the accuracy of the user-fitted 3DMMs, the time that it took to fit the model, as well as the overall subjective experience of users.  Since we cannot obtain the outputs of BodyTalk in the first study, we asked humans to rate the results, and determine which of them fits more accurately to the input shape. We had 10 users perform both studies (5 males, 5 females aged between 23 to 54), 7 of which are novices and 3 of which were professional 3D modelers. We present the quantitative results of these studies in Table~\ref{user_study_results}. As can be seen, Semantify achieves better accuracy in less time. Note that theoretically only the baseline can reach zero error, but because it is very difficult to handle, the error obtained was higher than Semantify. In addition, we asked the users for feedback on the experience of using the different applications (A was ours, B was the alternative). Here are some examples of such reviews (more examples could be found in the supplemental file):

``Using application B, occasionally when changing one slider it affects the other, which causes the user to start all over again''. 

``Using application B every minor change in a certain slider generated major changes in the rest of the sliders''

``Application A was far friendlier and I sensed as though I maintained much more control over the different body features''

``The abundance of sliders on application B only made it harder to control, not the other way around''.

\begin{table*}[!ht]
\centering
\begin{tabular}{ll}
\begin{tabular}{lccc|c}
\toprule
& \multicolumn{3}{c}{Time (min)} \\
\cmidrule(lr){2-4}
& Males & Females & Total & Score\\
\midrule
BodyTalk & 3.22 & 3.16 & 3.19 & 0.04\% \\
ours & \textbf{2.12} & \textbf{1.45} & \textbf{1.59} & \textbf{0.96\%}\\
\bottomrule
\end{tabular}
\hspace*{0.5 cm}
\centering
\begin{tabular}{lccc|ccc}
\toprule
& \multicolumn{3}{c}{Mean Error (cm)} & \multicolumn{3}{c}{Time (min)} \\
\cmidrule(lr){2-4} \cmidrule(lr){5-7}
& Males & Females & Total & Males & Females & Total \\
\midrule
Baseline & 0.122 & 0.111 & 0.116 & 3.18 & \textbf{2.09} & 2.43 \\
ours & \textbf{0.121} & \textbf{0.099} & \textbf{0.110} & \textbf{2.22} & 2.17 & \textbf{2.19} \\
\bottomrule
\newline
\end{tabular}
\end{tabular}
\caption{
User-study results for the SMPL 3DMM compared against BodyTalk~\cite{Streuber:SIGGRAPH:2016} (left, Score means the percent that this model was selected as better fitting the input), and SMPL-X parametric model compared against the raw blendshapes (right).  
}
\label{user_study_results}
\end{table*}

\subsection{Zero Shot Reconstruction}
\label{ssc:zeroShot}

To measure our zero-shot 3D-shape reconstruction method of SMPLX model, we use HBW (Human Bodies in-the-wild) dataset presented in SHAPY~\cite{Shapy:CVPR:2022}, containing  ground-truth 3D body scans with in-the-wild images. Table \ref{tab:zeroShot} shows the performance of our method compared to SHAPY and PIXIE~\cite{PIXIE:3DV:2021} on all of HBW validation set. Note that SHAPY also takes keypoints (usually obtained from openPose~\cite{Cao2018OpenPoseRM}) as input. Some examples are shown in Figure~\ref{fig:image2shape}. As can be seen, our zero-shot performance are on-par with state of the art methods, but we have the advantage of using semantic descriptors. This allows the user to fine tune the results from a very good initial guess, simply by using the semantic sliders. Some examples results with around a minute of fine-tuning can be found in the supplemental file.

\begin{table}
\centering
\begin{tabular}{lccc}
\toprule
& Males & Females & Total \\
\midrule
 SHAPY & \textbf{0.0196934} & \textbf{0.014422} & \textbf{0.017085} \\
   PIXIE & 0.024779 & 0.015265 & 0.020324 \\ 
   
Semantify (ours) & 0.022632 & 0.016564 & 0.019666  \\
\bottomrule
\newline
\end{tabular}
    \caption{Comparing our zero-shot shape reconstruction from image method against SHAPY\cite{Shapy:CVPR:2022} and PIXIE \cite{PIXIE:3DV:2021}. The values demonstrate the average MSE over HBW's in-the-wild validation set.}
        \label{tab:zeroShot}
\end{table}

\section{Conclusion}
\label{sec:conclusion}

We have presented a method to edit a 3D parametric model using semantically meaningful descriptors by first choosing a subset of descriptors judiciously and then learning a non-linear mapping from this set to the 3DMM coefficients. All this is done in a self-supervised manner harnessing the abilities of CLIP visual-language model. Our method can support a more intuitive user interface that allows even novices to model 3D shapes. It also supports other applications such as zero-shot reconstruction and fine-tuning of shapes from images.

\paragraph{Limitations and future directions.}
There are several limitations to our method.
First, our mapper's performance relies heavily on the data that is created (Section~\ref{ssc:data-creation}). For example, a dataset must contain extreme samples to achieve a more expressive mapper. Second, our goal with Semantify was to build a semantic representation of a given 3DMM without a human-in-the-loop. However, in some cases manually tuned modeling for a specific 3DMM would probably improve the 3DMM mapper's performance (we further elaborate on this in the supplemental file). Lastly, as shown in many other works, CLIP has rich semantic understanding, but there are cases in which its performance degrades. For example, performance is better with textured meshes, so using various textures for each 3DMM may improve the mapper's performance. Furthermore, more subjective descriptors such as ``cute'', ``scary'', ``pretty'' etc., are harder to rate using CLIP, and would probably not work well in our model.  

Future work can also improve the zero-shot 3D shape reconstruction by using prior information about its context (for example, the gender of a person, body keypoints as used in \cite{Shapy:CVPR:2022} etc.). Adding such priors  along with CLIP's semantic understanding may surpass the existing state-of-the-art solutions. 

\section{Acknowledgments}
This research was supported by the Israel Science Foundation (grant No. 1390/19).
We would also like to thank Eyal Gomel for valuable discussions.

\bibliographystyle{ieee_fullname}
\bibliography{clip2mesh}

\clearpage

\section{Supplementary Material}
\subsection{Background}
\label{sec:background}
In this paper, we focus our experiments on four main 3D Morphable Models and their variants. 
FLAME~\cite{FLAME:SiggraphAsia2017} is a 3DMM for human heads which consists of identity and expression spaces, using $N=5023$ vertices along with 4 joints. Similarly, SMPL and SMPLX~\cite{SMPL-X:2019} model human bodies using shape and expression spaces, with 23 joints and $N=6890$ vertices (SMPL), or 54 joints and $N=10,475$ and vertices (SMPLX). SMAL~\cite{Zuffi:CVPR:2017} was constructed using 3D scans of toy animals, and can represent a continuous space of animal shapes.

\subsubsection{FLAME}
FLAME uses standard vertex-based LBS with corrective shape, with N=5023 vertices and K=4 joints and is described by a function ${M}$ that returns ${N}$ vertices: 
\begin{align*}
        {M}(\vec{\beta},\vec{\theta},\vec{\psi}):\mathbb{R}^{\vec{|\beta|}\times\vec{|\theta|}\times\vec{|\psi|}} \rightarrow \mathbb{R}^{3N} \\ \forall \vec{\beta}\in\mathbb{R}^{\vec{|\beta|}}, \vec{\theta}\in\mathbb{R}^{\vec{|\theta|}}, \vec{\psi}\in\mathbb{R}^{\vec{|\psi|}}
\end{align*}
Where ${\beta}, {\theta}, {\psi}$ represent the coefficients of the shape, pose and expression respectively. The variations in the shape of different subjects are modeled by linear blendshapes as: 
\begin{equation}\label{first}
        B_{S}(\vec{\beta};{S}) = \sum_{n=1}^{\vec{|\beta|}}{\beta}_{n}{S}_{n}
\end{equation}

where ${\vec{\beta}} = [{\beta}_{1}, ..., {\beta}_{\vec{|\beta|}}]^{T}$ denotes the shape coefficients and ${S} = [{S}_{1}, ..., {S}_{\vec{|\beta|}}] \in \mathbb{R}^{3N\times\vec{|\beta|}}$ denotes the orthonormal shape basis. Similarly, the expression blendshapes are modeled by linear blendshapes as
\begin{equation}\label{second}
        B_{E}(\vec{\psi};{\varepsilon}) = \sum_{n=1}^{\vec{|\psi|}}{\psi}_{n}{E}_{n}
\end{equation}

where ${\vec{\psi}} = [{\psi}_{1}, ..., {\psi}_{\vec{|\psi|}}]^{T}$ denotes the expressions coefficients, and ${\varepsilon}=[{E}_{1}, ..., {E}_{\vec{|\psi|}}] \in \mathbb{R}^{3N\times\vec{|\psi|}}$ denotes the orthonormal expression basis. In this paper we use the first 10 principal components of the shape $\vec{\beta}$ and the first 10 principal components of the expression ${\vec{\psi}}$

\subsubsection{SMPL/SMPL-X}
SMPL-X stands for SMPL eXpressive, with shape parameters trained jointly for the face, hands and body. Similarly to FLAME, SMPL-X uses standard vertex-based LBS with learned corrective blendshapes, with N=10,475 vertices and K=54 joints and is described by a function ${M}$ that returns ${N}$ vertices:
\begin{align*}
    {M}(\vec{\beta},\vec{\theta},\vec{\psi}):\mathbb{R}^{\vec{|\beta|}\times\vec{|\theta|}\times\vec{|\psi|}} \rightarrow \mathbb{R}^{3N} \\ \forall \vec{\beta}\in\mathbb{R}^{\vec{|\beta|}}, \vec{\theta}\in\mathbb{R}^{3({K}+1)}, \vec{\psi}\in\mathbb{R}^{\vec{|\psi|}}
\end{align*}
Where ${\beta}, {\theta}, {\psi}$ represent the coefficients of the shape, pose and expression respectively, and the shape blendshapes function is the same as \eqref{first}. In this paper, we use the first 10 principal components of the shape ${\vec{\beta}}$.

\subsubsection{SMAL}
Analogous to SMPL, the SMAL function is also defined by ${M}(\beta,\theta,\gamma)$ such that $\beta,\theta,\gamma$ represent the shape, pose and translation respectively.
$\beta$ is a vector of the coefficients of the learned PCA shape space, $\theta \in \mathbb{R}^{3N}=\{r_{i}\}_{i=1}^{{N}}$ is the relative rotation of the N=33 joints in the kinematic tree, and $\gamma$ is the global translation applied to the root joint. The SMAL function returns 3D mesh's vertices, where the template model is shaped by $\beta$, articulated by $\theta$ through LBS, and shifted by $\gamma$. In this paper, we use the first 10 principal components of the shape $\vec{\beta}$.

\subsection{Word Descriptors}
We provide our initial sets of descriptors for each one of the models. Colored words represent the final set of descriptors that were chosen by our method.

\subsubsection{Body}
The color coding is blue for \color{blue}SMPLX\color{black}, red for \color{red}SMPL\color{black}, and cyan when the word was chosen for \color{cyan}Both \color{black}  models.
\paragraph{\color{black}Male model:}
short, tall, \color{red}long legs\color{black}, big, \color{cyan}fat\color{black}, broad shoulders, built, curvy, fit, heavyset, lean, \color{blue}long torso\color{black}, long, \color{blue}muscular\color{black}, pear shaped, \color{blue}petite\color{black}, \color{red}proportioned\color{black}, \color{blue}rectangular\color{black}, round apple, short legs, short torso, skinny, \color{blue}small\color{black}, stocky, strudy, \color{red}narrow waist\color{black}, thin.
\paragraph{\color{black}Female model:}
\color{cyan}fat\color{black}, thin, hourglass, short, \color{red}long legs\color{black}, \color{cyan}narrow waist\color{black}, skinny, tall, broad shoulders, pear shaped, average, big, curvy, lean, \color{cyan}proportioned\color{black}, sexy, fit, heavyset, \color{blue}petite\color{black}, \color{red}small\color{black}.
\paragraph{\color{black}Neutral model:}
short, \color{blue}tall\color{black}, long legs, big, \color{cyan}fat\color{black}, curvy, feminine, fit, heavyset, lean, \color{cyan}long torso\color{black}, long, \color{cyan}masculine\color{black}, muscular, pear shaped, \color{red}petite\color{black}, \color{blue}proportioned\color{black}, \color{blue}rectangular\color{black}, round apple, short legs, short torso, skinny, \color{cyan}small\color{black}, stocky, strudy, attractive, sexy, \color{red}narrow waist\color{black}, \color{blue}hourglass\color{black}.

\subsubsection{Face Model}
The color coding is \color{cyan}cyan \color{black} for the chosen descriptors.
\paragraph{\color{black}Shape:}
\color{cyan}fat\color{black}, thin, \color{cyan}long neck\color{black}, \color{cyan}big forehead\color{black}, nose sticking-out, \color{cyan}ears sticking-out\color{black}, \color{cyan}small chin\color{black}, long head, chubby cheeks, big head.
\paragraph{Expression:}
happy, sad, angry, surprised, disgusted, fearful, neutral, \color{cyan}smiling\color{black}, \color{cyan}serious\color{black}, pensive, confused, bored, sleepy, tired, excited, relaxed, calm, nervous, worried, scared, \color{cyan}open mouth\color{black}, raise eyebrows, open eyes, smile.

\subsubsection{Animals Model}
The color coding is \color{cyan}cyan \color{black} for the chosen descriptors.

\paragraph{Shape:} hippo, donkey, horse, \color{cyan}cow\color{black}, \color{cyan}lion\color{black},\color{cyan}cat\color{black}, \color{cyan}dog\color{black}.

\begin{table*}
\color{black}
\centering
%\resizebox{\textwidth}{!}{%
\begin{tabular}{lccccc|ccc}
\toprule
& \multicolumn{5}{c|}{number of descriptors} & \multicolumn{3}{c}{number of samples}\\
\cmidrule(lr){2-6} \cmidrule(lr){6-9}
& 2 & 5 & 6\color{red}* & 10 & 15 & 1K & 3K\color{red}* & 10K\\
\midrule
Error (cm) & 0.0372 & 0.0219 & 0.0233 & 0.0087 & 0.0047 & 0.0247 & 0.0233 & 0.0154 \\
Steps & 568 & 2402 & 2684 & 3769 & 4679 & 2932 & 2684 & 4034 \\
\bottomrule
\\
\end{tabular}%

%}
\caption{
Model expressiveness ablation study. These are ablations of different mappers that were trained on various numbers of descriptors and different numbers of samples.
%and with and without texture. 
Columns with \color{red}* \color{black} represent our final configurations. Steps indicate the average number of steps that took the optimization process to converge. The error is evaluated on a scale of cm. Although 15 descriptors results in the lowest error, the semantic meaning of the descriptors degrades due to correlations between the descriptors. 
}
\label{smplx_comparison}
\end{table*}

\subsection{CLIP-Based Optimization}
\begin{figure}[ht!]
\begin{center}
\includegraphics[width=0.5\textwidth]{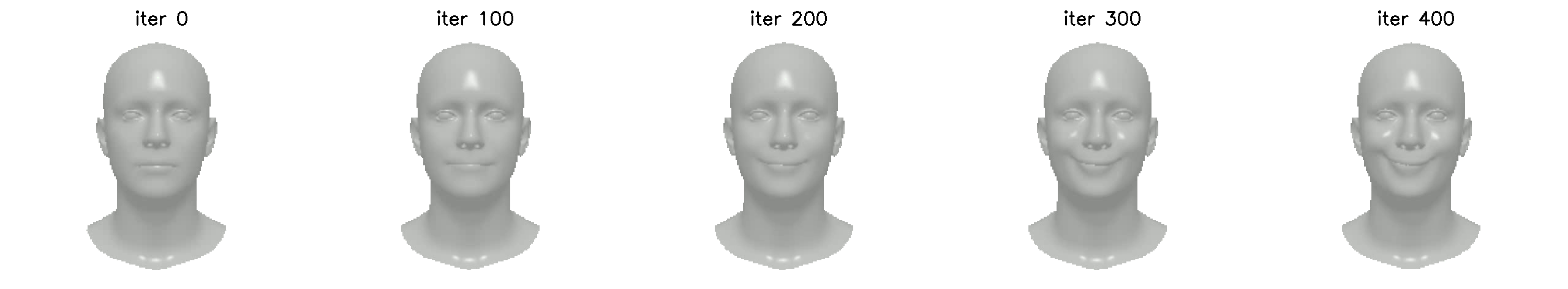}
    \caption{CLIP-based optimization visualization by iteration}
    \label{clip_optimization}
\end{center}
\end{figure}

To estimate how a single descriptor will affect the 3DMM with respect to CLIP's semantic understanding, we ran CLIP-based optimization experiments. Figure \ref{clip_optimization} demonstrates the optimization's results with respect to ``smile'' descriptor on FLAME model. This method was not a good estimation for the effect of the descriptors for two reasons:
\begin{enumerate}
    \item In our method CLIP is used for rating (that is, we use CLIP's scores for a given image that has already been deformed and a given set of descriptors), rather than using CLIP as to edit the mesh using its semantic understanding of a given set of descriptors.
    \item CLIP-based optimization optimizes a single descriptor each time (feeding multiple descriptors together may enforce putting them in a sentence to optimize CLIP's performance), therefore the relations that appear between descriptors in our model could not be foreseen by using this method.
\end{enumerate}

\subsection{Clustering}
\begin{figure}[ht!]
\begin{center}
\includegraphics[width=0.5\textwidth]{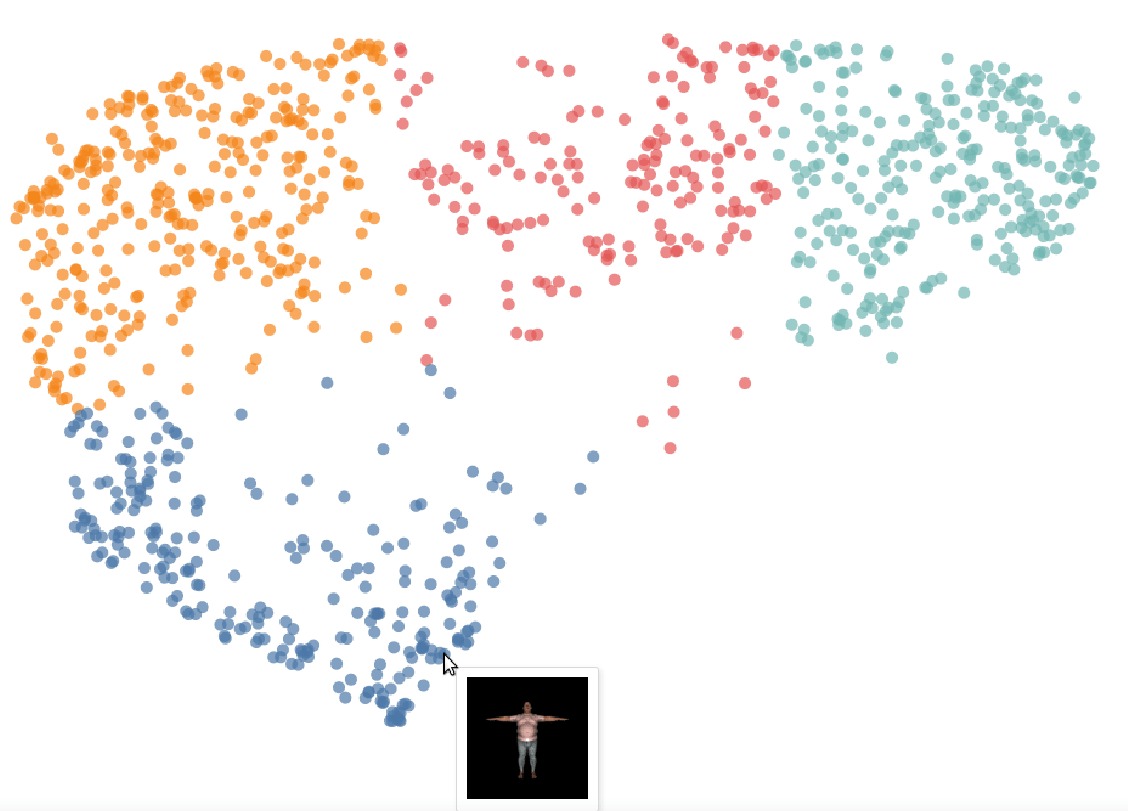}
     \caption{Clustering the encoded images after reducing the dimension by U-Map}
     \label{umap}
\end{center}
\end{figure}

Prior to clustering the images in CLIP embedding space's dimension, we used Uniform Manifold Approximation and Projection (U-MAP) to reduce the dimension of the data in order to visualize it and verify that close images resemble each other and distant images differ (an example for such visualization could be found in Figure~\ref{umap}).

\begin{figure*}
\begin{center}
\includegraphics[width=0.95\textwidth]{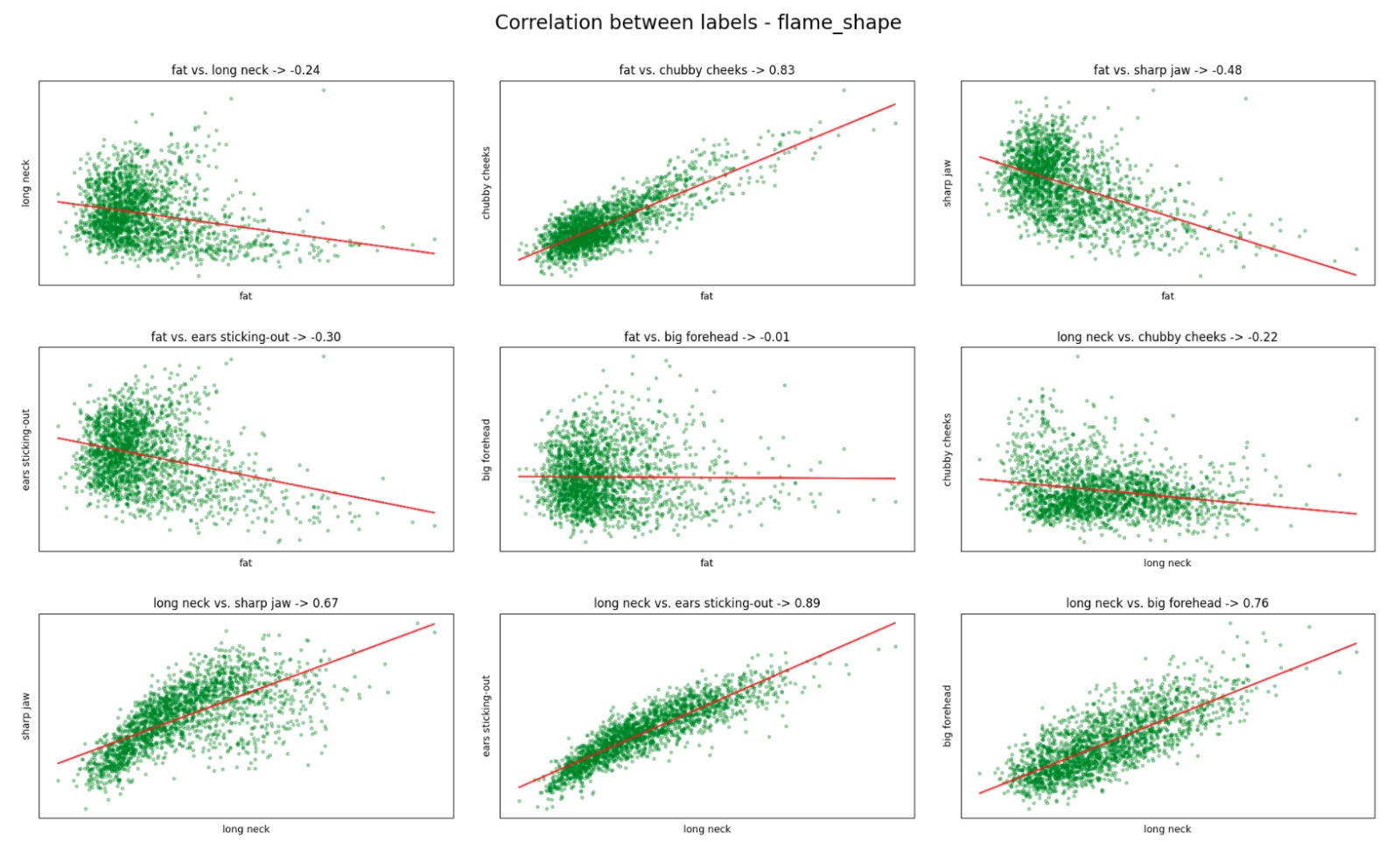}
    \caption{Example for the correlation between shape descriptors (in this case from FLAME 3DMM).}
    \label{flame_shape_correlation}
\end{center}
\end{figure*}

\subsection{Manually Tuned Model}
\label{manually_model}
Our goal in this paper was to present a novel general method for mapping the semantic representation to its parametric counterpart \emph{without human-in-the-loop}.
As noted in our conclusions, our method relies mainly on the data that is generated to train the Mapper, therefore, outliers in the data (which in 3D meshes corresponds with ``broken'' meshes) would probably lead to a degradation in the Mapper's performance. A case-specific solution might produce a better performance since images could be created manually or in a more supervised and case-specific manner. Some examples of such an implementation:
\begin{itemize}
    \item When using SMAL model, by sampling instances randomly, it is hard to generate instances from the Hippopotamidae family (without ``breaking'' the mesh), whereas generating it non-randomly is quite an easy task.
    \item For the zero-shot shape prediction from an image, it is possible to fine-tune the Mapper by feeding extreme instances or generating specific types of body shapes that would improve the predictions.
\end{itemize}

\begin{figure*}
\begin{center}
\includegraphics[width=0.95\textwidth]{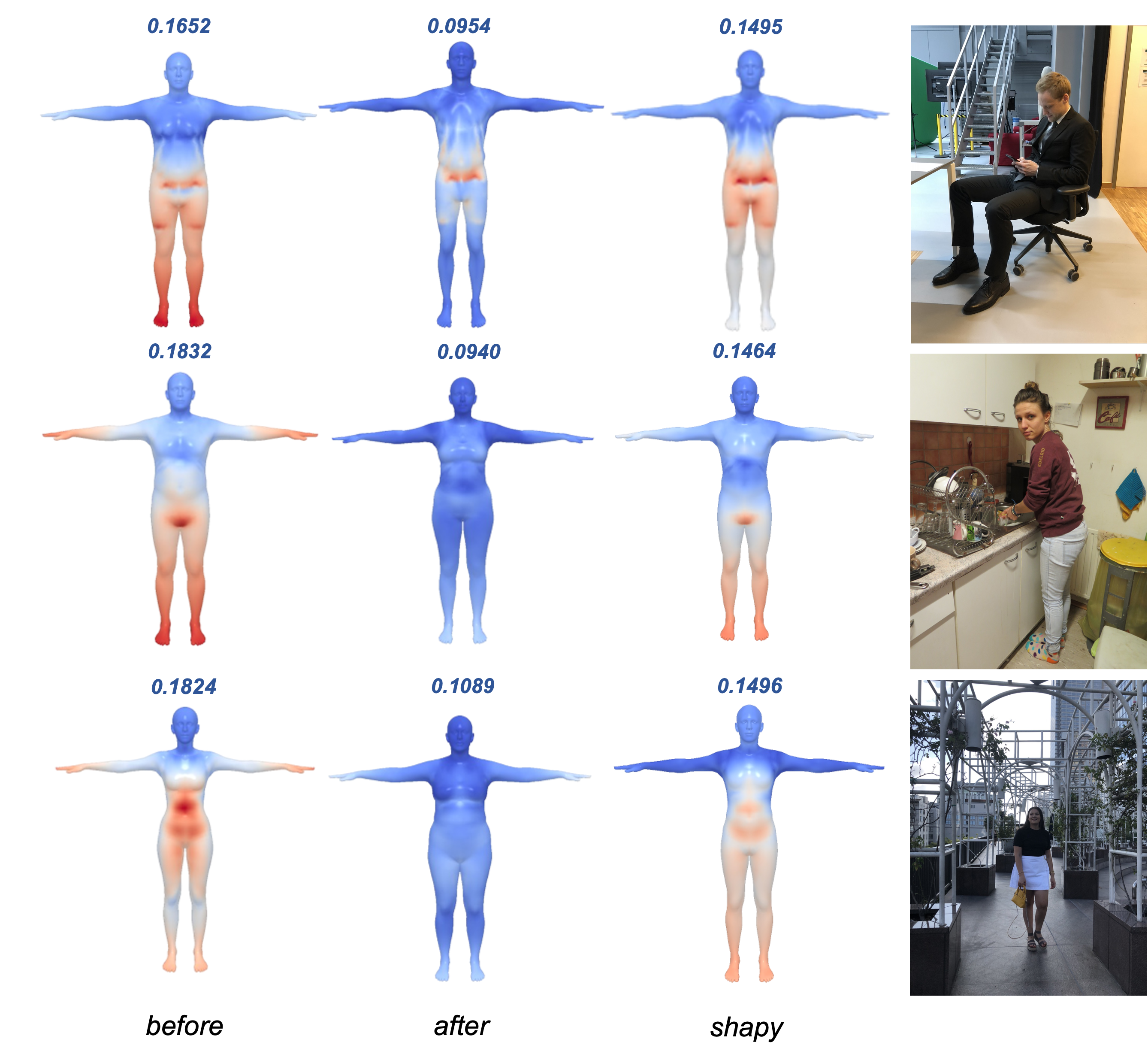}
    \caption{Zero-Shot Image to Shape Reconstruction task with fine-tuning the predicted shape with respect to the image using our interactive application. The initial prediction is on the left, then the fine-tuned shape and on the right is SHAPY's prediction. Fine tuning takes around 1 minute.}
    \label{fig:finetune}
\end{center}
\end{figure*}

In addition, using Semantify as zero-shot predictor of shape from images allows the user to easily fine-tune the results and reach a better fit as can be seen in the examples in Figure~\ref{fig:finetune} where fine-tuning of around one minute was applied to the zero shot results.

\subsection{User Study - Feedbacks}
We asked the users to provide us with feedback on their experience of using both applications as a tool to fit a 3D shape (A was ours, B was the alternative). Here are more examples of such reviews:
\begin{itemize}
    \item When using application B, words are sometimes not clear or not relevant, while on the other hand, clear and straightforward words such as ``fat'' and ``thin'' are missing.
    \item Application B's sliders are not ``frozen'', so occasionally when changing one slider it affects the other, which causes the user to start all over again.
%    \item Application B is not user-friendly in terms of rotating the figures or switching the point of view.
    \item In terms of user experience using the different sliders, application A was far more comfortable to use than B. It was clearer what was meant to be the effect of each slider, and each change influenced that specific body feature alone. Conversely, application B every minor change in a certain slider generated major changes in the rest of the sliders in a way that made it more difficult to control the result. The abundance of sliders on application B only made it harder to control, not the other way around. Ultimately, application B created a figure which was less muscular, and paid little attention to detail in regards to the body curves and body fat. Overall, the results were undeniably better in application A.
    \item Trying out the application B ... even the slightest touch of a single slider that was intended to get me closer to my desired goal, prompted a significant change in 10 different sliders which completely and utterly ruined what I was aiming for. To the contrary, the experience with application A was far friendlier and I sensed as though I maintained much more control over the different body features. With regards to the final results, you just can’t draw a comparison. Juxtaposing the two finalised models makes it humorously obvious that application A wins by a landslide.
\end{itemize}

% \bibliographystyle{ieee_fullname}
% \bibliography{clip2mesh}

\begin{table}[h]
\color{black}
\centering
\begin{tabular}{lccccc}
\toprule
\multicolumn{6}{c}{Number of samples used for training the model}\\
& 1k & 3k\color{red}* & 7k& 10k\\
\midrule
Error (cm) & 0.0026 & 0.0027 & 0.0022 & 0.0025 & \\
Steps & 2850 & 1203 & 4618 & 2328 & \\
\end{tabular}%
\\
\begin{tabular}{lccccc}
\toprule
\multicolumn{6}{c}{Number of descriptors used in the model}\\
& 2 & 5 & 6\color{red}* & 10 & 15\\
\midrule
Error (cm) & 0.003 & 0.0028 & 0.0027 & 0.0023 & 0.002 \\
Steps & 1813 & 1466 & 1203 & 4795 & 4999\\
\bottomrule
\end{tabular}%
\caption{Ablation results on 10 meshes that were registered to range scans  from CAESER dataset, we used chamfer distance as an objective to fit our sementified sliders with our pretrained mapper to the GT mesh. The  real-world captured shapes from CAESER dataset on our SMPL-X \textbf{male} mapper.}
\label{tab:ablation}
\end{table}

\end{document}